\documentclass[onecolumn]{IEEEtran}
\usepackage{makeidx}
\usepackage{multirow}
\usepackage{multicol}
\usepackage[dvipsnames,svgnames,table]{xcolor}
\usepackage{graphicx}
\usepackage{epstopdf}
\usepackage{amsmath}
\usepackage{amssymb}
\usepackage{csquotes}
\usepackage{color}
\usepackage{mathtools}
\usepackage{algorithm}
\usepackage{algorithmic}
\usepackage[normalem]{ulem}
\usepackage{verbatim}

\usepackage[utf8]{inputenc}

\newcommand{\bt}{{\bf t}}
\newcommand{\by}{{\bf y}}
\newcommand{\bx}{{\bf x}}
\newcommand{\bw}{{\bf w}}

\newcommand{\bb}{{\bf b}}

\newcommand{\bA}{{\bf A}}

\newcommand{\bL}{{\bf L}}
\newcommand{\bC}{{\bf C}}
\newcommand{\bI}{{\bf I}}

\newcommand{\bT}{{\bf T}}

\newcommand{\bX}{{\bf X}}

\newcommand{\bdp}{{\bf dp}}
\newcommand{\btemp}{{\bf temp}}

\newcommand{\bu}{{\bf u}}

\newcommand{\bmu}{{\boldsymbol \mu}}
\newcommand{\bSigma}{{\boldsymbol \Sigma}}
\newcommand{\bsigma}{{\boldsymbol \sigma}}

\usepackage{hyperref} 

\title{Data Selection for Short Term load forecasting} 
\author{Nestor Pereira, Miguel Ángel Hombrados Herrera, Vanesssa Gómez-Verdejo, Andrea A. Mammoli, Manel Martínez-Ramón}

\begin{document}

\maketitle


\begin{abstract}
\noindent Power load forecast with Machine Learning is a fairly mature application of artificial intelligence and it is indispensable in operation, control and planning. Data selection techniqies have been hardly used in this application. However, the use of such techniques could be beneficial provided the assumption that the data is identically distributed is clearly not true in load forecasting, but it is cyclostationary. In this work we present a fully automatic methodology to determine what are the most adequate data to train a predictor which is based on a full Bayesian  probabilistic model. We assess the performance of the method with experiments based on real publicly available data recorded from several years in the United States of America. 
\end{abstract}

\section{Introduction}

Power load forecasting has  been an indispensable tool in the operation, control and planning of the power grid over  decades. On the daily operation of a power grid,  load forecasting is required to guarantee that energy generated exactly matches power demand. 
It is essential on operation  tasks like unit commitment or hydro plants scheduling, but also is required on estimating the distribution lines loading and energy trading. In the last years, due to the deregulation of the power market and the beginning of a shift towards a new paradigm of power grid --the so called Smart Grid and its variants \cite{EPRI2016}, forecasting of electric parameters in general, and power load in particular, have become even more relevant \cite{Hong:2016}.

There are many studies that propose methods for grouping together load patterns with similar curve shapes, this is usually known as load profiling on the context of power load forecasting \cite{panapakidis2016application}. In many cases, this procedure simply consists of 
dividing  manually the data set into smaller subsets based on previous knowledge of the particular data. A very common technique, shown in \cite{Charlton:2014},\cite{Mori:2005}, \cite{hsu1991artificial} and \cite{Feng:2016}  is to make a selection based on a particular day, that is, training data are selected of the same nature as the sample is going to be predicted, i,e:  weekdays, weekends or holidays. Another common case, also present in these papers, is to split the samples according to seasons and generate a different model for each season. Few papers, such as \cite{Zivanovic:2001}, have selected samples from the original set that match specific requirements with tomorrow's load model like similar maximal and minimal temperatures and calendar dates. 

Machine learning provides countless tools for clustering and automatic selection of samples. These techniques allow the  automatic selection of the most informative samples without the need of making previous assumptions on the characteristics of the data. An example of this is \cite{nagi2011computational}, where a self-organizing map (SOM) algorithm is used for clustering samples into two subsets according to a similarity criterion of the load data in order to predict the daily peak load of the next month with a support vector machine (SVM). In \cite{mori2001deterministic}, deterministic annealing clustering (DA) is used in order to classify the input data of an artificial neural network (ANN) that performs one-step ahead daily maximum load forecasting. In some cases, the methodology for clustering and forecasting only includes a clustering method but not a conventional supervised phase for  conventional forecasting \cite{panapakidis2016application}. For example,  in \cite{lopez2012application}, a SOM is used for clustering the data, then
the target day is associated with the  most similar data by means of the known load data available from the target day. The forecasted load is selected within the samples selected by similarity in order to predict the load profile of sessions of the daily and intra-daily Spanish market 

A common fashion to categorize load forecasting is based on the prediction horizon:  Short term load forecasting (STLF), usually from a day up to two weeks, medium term load forecasting (MTLF), from two weeks to three years and long term load forecasting (LTLF), several years.
One of the  most common time frame on the short term power load forecasting literature is hourly or half-hourly day-ahead forecasting (\cite{zhang2015day,taylor2007short,sahay2014day}). In the latter, for example, authors discuss the performance of artificial intelligence (AI) on STLF, in particular, on hourly day ahead load forecasting.

Load forecasting techniques can be found in the literature that use statistical techniques and artificial intelligence (AI).  We can include among the former approaches  auto regressive and moving average (ARMA) models, multilinear regressor models  (MRL) or exponential smoothing models. Nevertheless, the line that divide them is unclear due to the combination of multiple disciplines on the research community. For example, ARMA models can be optimized using ML approaches \cite{Rojo2004}, and they can be generalized using kernel methods \cite{Martinez2006,Rojo2018}. Review paper  \cite{abu1982short} makes a review of these and other techniques for its application to STLF. 

Algorithms that use AI can be taxonomized in different groups depending on their linear or nonlinear nature, their structure, that can be single layer or multilayer of the criteria used for their optimization. Common criteria include minimization of the mean square error (MMSE), maximization of the classification or regression margin, maximum likelihood (ML) or maximum a posteriori (MAP) for the case of Bayesian machine learning approaches \cite{Murphy2012}.

For example, a Least Square Support Vector Machine LS-SVM is used in \cite{yang2011comparison}, where the used solution is a maximum margin \cite{Vapnik1995,Suykens1999} linear algorithm; nonlinear algorithms that use the kernel trick \cite{Shawe2004} are  kernelized versions of Support Vector Regressor \cite{smola2004} or kernel Gaussian processes \cite{Rasmussen:2006}, which use Bayesian criteria for optimization. Nonlinear  SVMs  are used in  \cite{chang2008support}, and Gaussian processes in  \cite{yan2012load}.  Multilayer models based upon artificial neural networks \cite{Bishop1995}  are presented in   (\cite{panapakidis2016application},\cite{rana2016forecasting})  for load forecasting. Alternative models that are not classified in either of the above mentioned  categories are fuzzy methods, used in \cite{coelho2016self} or \cite{ye2006identification}.

Another form of classifying the forecasting methods is between  point load forecasting  and   probabilistic load forecasting (PLF). As opposed to the former, PLF algorithms provide a prediction interval that quantifies the uncertainty of the prediction. Although the number of publications of probabilistic methods for load forecasting is a minority with respect traditional methods, PLF are becoming more relevant since are of extreme value in tasks like prediction of equiment failure and the integration of renewable energy into the grid \cite{Hong:2016}. Gaussian Process (GP) is one of the most popular algorithms for forecasting prediction in general  and for PLF in particular. Some examples of works that used GP in the context of energy and power are  \cite{kou2014sparse} or \cite{mori2005probabilistic}.

There is a generalization of the GP that allows the prediction of multiple outputs simultaneously, this type of algorithm is usually known as multitask GP or multioutput GP. The advantage of this method over the single output GP is that the multitask GP is able to exploit the potential correlations between the output variables in order to improve its performance. 

The multioutput prediction is specially advantageous, for example,  when there are missing  data \cite{borchani2015}. In the last years, many approaches for multitask GP have been presented, \cite{rakitsch2013all, stegle2011efficient,bonilla2008multi,boyle2005multiple, Boyle05dependentgaussian}. In  \cite{zhang2014power}, for example, a multitask GP is used to perform load forecasting of several cities simultaneously.

Most of the works focused on load forecasting use as explanatory variables past values of the load as well as as weather variables since weather, in particular temperature,  is a major factor affecting the electricity demand \cite{Hong:2016}. Works like \cite{peirson1994electricity} or \cite{hong2015weather} evaluate the correlation of weather variables with power load and its importance as explanatory variables for improving load prediction.

This paper presents a new method for data selection that, together with a Gaussian Process Regressor, is aimed to perform day-ahead predictions, therefore it can be classified as a short term forecasting. The time horizon is  hourly day-ahead load forecasting at aggregated levels. While a significant number of the works present in the power load forecasting literature use heuristic methods applied \textit{ad-hoc} for each particular  data set and predicting time horizon, the method proposed is fully automated. This method is based on a Bayesian approach and does not require any prior knowledge about the data in order to select the samples for the forecasting algorithm. The selection method allows a more compact and less complex prediction algorithm since it dramatically reduces the number of samples required for day ahead prediction. As a consequence of that, the algorithm is not intended for generating single annual  model, but  for being trained before every daily prediction. This may seem a priory computationally expensive, but in our experiments the computation time is affordable and it is justified by a significant improvement in the prediction performance. 

\section{Bayesian data selection for training dataset construction}

A possible approach in short term load forecast is to consider a long dataset consisting of samples across as many years as possible. Then, a machine learning algorithm can be trained in order to capture the necessary information to forecast the load at any time of the year, including all seasons. When using kernel approaches as Gaussian Processes, this may lead to a large matrix that will be hard to invert. Also, the variability across the year may be not well represented. An alternative could be constructed that trains a compact machine adapted to a given time instant, then producing a single prediction. After that, the set of parameters can be discarded if the training time is negligible. If the training time is not negligible, in some cases an incremental training can be applied for the next time instant. In any case, the training dataset must be chosen so it captures the probabilistic properties of the predictor or test input  sample to be evaluated. 

The strategy presented in this work consists of estimating a posterior distribution of the time given the observation, in order to pick the most probable time instants from this distribution. In other words, given a large dataset with samples conveniently labelled with their corresponding time of the year instants and  a single test predictor $\bx_n$, the question to be answered is what is the probability of that the test  sample belongs to a given time of the year different from its actual one. If this probability is high, we will pick as training pattern the sample that actually belongs to that time of the year as a training sample.  

In order to formalize this idea, consider a set of observations $\bx_n$, $1{\leq}n{\leq}N$, $\bx_n \in\mathbb{R}^{D}$ and where each sample has a time stamp associated $1{\leq}{t}_n{\leq}N$, $t_n  \in\mathbb{R}$ .
More than one sample can have the same time stamp. For example $t_n$ indexes can represent each day of the year; thus, two observations $\bx_n$ and $\bx_m$ of the same day in different years will have equal timestamps $t_n=t_m$.

\subsection{Probabilistic model}
Assume that $t_n$ has a posterior probability distribution $p(t_n | \bx_n)$ with respect its associated observation $\bx_n$. By virtue of the Bayes' rule, this posterior can be computed as 

\begin{equation}\label{eq:posterior}
p(t_n | \bx_n) = \frac{p(\bx_n | t_n)\cdot p(t_n)}{p(\bx_n)}
\end{equation}

Where $p(t_n)$ is the prior  probability of time $t_n$. This, being interpreted as the probability of picking a time instant at random, may be modelled as a uniform variable, hence a constant $p(t_n)=T^{-1}$, whwre $T$ is the number of samples available for a year period. Thus, the following proportionality expression holds true 
\begin{equation}\label{eq:prop_posterior1}
p(t_n | \bx_n) {\propto}  \frac{p(\bx_n | t_n)}{p(\bx_n)}
\end{equation}

Also, since the marginal likelihood $p(\bx_n)$ is independent of $t_n$, we can also see that the posterior of $t_n$ is proportional to the conditional likelihood of $\bx_n$, i. e.  
\begin{equation}\label{eq:prop_posterior2}
p(t_n | \bx_n) {\propto}p(\bx_n | t_n)
\end{equation}
If this likelihood is modelled as a normal distribution $p(\bx_n | t_n) = \mathcal{N}(\bx_n|\bmu_{1:D},\bSigma_{1:D})$, then we can say that 
\begin{equation}
    p(t_n|\bx_n)\propto \mathcal{N}(\bx_n|\bmu_{1:D},\bSigma_{1:D})
\end{equation}
where $\bmu_{n,1:D}$ and $\bsigma_{1:D}$ are the mean and the covariance matrix of the distribution of the $D$-dimensional variable $\bx_n$. By using the chain rule of probability, the likelihood can be factorized as 
\begin{equation}\label{eq:MTGP_factorized}
\begin{split}
&p(\bx_{n}| t_n)=\\
&=p(x_{n,D}|\bx_{n,1:D-1}, t_n)\cdot p(x_{n,D-1}|\bx_{n,1:D-2}, t_n)\cdots\\
&\cdots p(x_{n,2}|, x_{n,1}, t_n) \cdot p(x_{n,1}|, t_n)
\end{split}
\end{equation}  
where $x_{n,d}$ is the $d$-th component of $\bx_n$ and $\bx_{n,1:d}$ is a vector containing the $d$ first components of it. It can also be assumed that since the likelihood is a Gaussian distribution, the factors in \eqref{eq:MTGP_factorized} are also  $D$ univariate Gaussian distributions.

At this point, each of these conditional  distributions are further modelled as univariate linear regression  Gaussian process models

\begin{equation}\label{eq:univariate_definitions}
\begin{split}
p(x_{n,d}|\bx_{n,1:d-1},t_n)&=\mathcal{N}(x_{n,d}|m_{n,d},\sigma_d^2)\\
m_{n,d}&=\bw_d^{\top} \cdot \bx_{n,1:d-1} + w_{t,d}^{\top} t_n
\end{split}
\end{equation}  
and where it is implicitly assumed that observation $x_{n,d}$ given $\bx_{n,1:d-1},t_n$ has a Gaussian observation error which is independent and identically distributed with variance $\sigma_d^2$. Therefore we assume that samples $x_{n,d}$ are conditionally independent given $\bx_{n:d-1}$ and $t_n$, with identical variance $\sigma_d^2$. Then, the joint likelihood for the process $\bx_d=\{x_{1,d}\cdots x_{N,d}\}$ can be written as
\begin{equation}\label{eq:joint_likelihood}
    p(\bx_d|\bX_{1:d-1},\bt)=\prod_{n=1}^N \mathcal{N}(x_{n,d}|m_{n,d},\sigma_d^2) =\mathcal{N}(\bx_d|\bw^{\top}\bX_{1:d-1}+w_{t,d}\bt,\sigma_d^2\bI)
\end{equation}
In this expression, matrix $\bX_{1:d-1}=[\bx_{1,1:d-1} \cdots \bx_{N,1:d-1}]$ contains all vectors $\bx_{n,1:d-1}$ of the training set, and $\bt$, defined here as a column vector, contains all timestamps $t_n$

By establishing a joint Gaussian prior for variables $\bw_d$ and $\bw_{\bt,d}$, with zero mean and covariance $\bSigma_{p,d}$, its posterior can be computed by the Bayes' rule. Then, the maximum a posteriori set of parameters is   \cite{Rasmussen:2006}

\begin{equation}\label{eq:mean_weights}
\left(
\begin{array}{c}
\bar{\bw}_d\\
\bar{w}_{t,d}
\end{array}
\right)=\sigma_d^{-2}\bA_d^{-1}\left(
\begin{array}{c}
\bX_{1:d-1}\\
\bt^{\top}
\end{array}\right)\bx_d
\end{equation}
where  $\bt$ is a column matrix containing all time stamps and matrix $\bA$  is defined as 
\begin{equation}\label{eq:covar_weights}
\bA_d=\sigma^{-2}_d\left(
\begin{array}{cc}
\bX_{1:d-1}\bX_{1:d-1}^{\top} &\bX_{1:d-1}\bt^{\top}\\
\bt\bX_{1:d-1}^{\top}&\sum_n t_{n}^2
\end{array}
\right)+\bSigma_{p,d}^{-1}
\end{equation}

The proof of these equations is given in  Appendix \ref{app:matrix_inversion}.

The above model computes the likelihood of sample $\bx_n$ given its associated  time instant only. Now let us assume a new sample $\bx^*$ at time instant $t^*$ that is to be used to predict a future magnitude with a predictor. We want to determine what other past time instants are likely to contain a this sample (or more specifically, a sample similar to $\bx^*$). 

The predictive posterior probability given in \cite{Rasmussen:2006} is derived by marginalizing the likelihood \eqref{eq:joint_likelihood} with respect to the posterior of parameters $\bw_d,w_t$, which is a Gaussian with mean and variance given by equations \eqref{eq:mean_weights} and \eqref{eq:covar_weights}. The result is a product of Gaussian distributions, each one with a mean  and a variance given by  
\begin{equation}\label{eq:posterior_variance}
\begin{split}
\mathbb{E}(x^*_{d})&=\bar{\bw}_d^{\top} \cdot \bx^*_{1:d-1} + \bar{w}_{t,d}^{\top} t_n\\
    \mathbb{V}ar(x^*_{d})&=\left[ \bx^{*\top}_{1:d-1},~ t_n\right]\bA_d^{-1} \left[ \bx^{*\top}_{1:d-1}, ~ t_n\right]^{\top}
   \end{split}
\end{equation}
with the mean values of the parameters defined in \eqref{eq:mean_weights}. NOte that here $t_n$ is any time instant. The corresponding proof is provided in Appendix \ref{app:posterior}.
\newline

It is important to remark that while the likelihood \eqref{eq:joint_likelihood} is modelled as a Gaussian function of $\bx^*$, the posterior in equations \eqref{eq:posterior} to \eqref{eq:prop_posterior2} is not a Gaussian as a function of $t_n$, and it is actually a multimodal distribution, as it will be shown in the experiments. Indeed, for a given test sample, if one sweeps the time across training samples, one expects that samples with a higher time posterior are these that belong to same or close days of past years, i.e. days of the same season, thus showing a cyclostationary component. 

\subsection{Data selection procedure}

Assume that a pattern $\bx_n$ is available to compute a given estimation. If this estimation is the forecast of a 24 hour ahead load, this pattern can contain information about the last measured power load, the present weather and the weather forecast, for example. The prediction model needs to be trained, and we choose to do it with data with characteristics that are similar to the predictor $\bx_n$.

The above probabilistic model is intended to choose time instants that contain data similar to is predictor. From an intuitive point of view, a suitable procedure would be to measure the euclidean distance between sample $\bx_n$ and all the training data and choose a subset of the closest samples for training. Alternatively, the above model computes what is the likelihood of sample $\bx^*$ given  every instant $t_m$, where $1 \leq m \leq N$. As it is been proven in previous subsection, this likelihood is proportional to the posterior of $t_m$ given $\bx_n$, and one can thus choose the data belonging to the time instants with a higher posterior probability.

Thus, the algorithm is very simple, and consists of the following steps:
\begin{enumerate}
    \item Given a training set $\{\bx_n\}_{n=1}^N$, compute the mean vector and covariance matrix of the parameter posterior in equations \eqref{eq:mean_weights} and \eqref{eq:covar_weights}. 
    \item Given a test sample $\bx^*$ Use equations \eqref{eq:univariate_definitions} and \eqref{eq:posterior_variance} to compute the mean and variance at every time instant. These statistics use the same value $\bx^*$, but all possible time instants $t_m$.  Compute the probability of every time instant using  \eqref{eq:MTGP_factorized}.
    \item Rank all data by its time posterior.     
\end{enumerate}

[SEE COMMENT FROM ANDREA. IS IT POSSIBLE TO IMPROVE THE EXPLANATION OF THE ALGORITHM?]

The data selection can be probabilistic if one chooses the data at random by drawing a sample at a time by using a random generator with a distribution equal to the obtained posterior. A deterministic procedure can be simply to choose a subset of the data that have the highest posterior values. This simple procedure is the one used in the experiments below.

\section{Experiments}

In order to test the performance of our selection algorithm, we compare three different data selection methods used to train the same Gaussian Process regression model. In all the experiments, the size of training set is set to the number of samples contained in two years. In order to analyze a week's temperature and load trend for a specific day $i$ to forecast, we study the hourly temperature and load values of the previous 7 days to the day to predict $X_{i-6:i}$. Since we use $i+1$ as a prediction based on day $i$, the last day of the years is also not used as an input training variable. For these reasons, only 357 samples out of the 365 of each year are used. The number of samples in all of our experiments is set to 714.

The three different data selection methods are the following: 
\begin{enumerate}
\item The first data selection method (closest time data, CT) consists of using 357 samples from 2016 and 357 samples from 2017, to construct  a training set of 714 samples including samples from all seasons, weekdays, weekend days and holidays.

\item The second method consists of selecting data by distance similarity, this is, we compute the euclidean distance between the test sample and all the training sample and choose the 714 samples closer to the test sample. We label this method as closest distance (CD) method. 

\item The third method uses the posterior probability distribution obtained with the Bayesian data selection algorithm presented in the previous section to select the most similar 714 samples to the day we want to predict. We call this method maximum a posteriori (MAP) method. 
\end{enumerate}

The model accuracy using the different data selection strategies for an individual prediction is compared using Mean Absolute Percentage Error (MAPE), defined as

\begin{equation}
    MAPE=\frac{100}{N_{test} \cdot D}\sum_{n=1}^{N_{test}} \sum_{k=1}^{D} \frac{|y_{n,k}-\hat{y}_{n,k}|}{y_{n,k}}
\end{equation}
where $D$ is the dimension of an individual $n$ target test sample, $N_{test}$ is the number of samples in the test set, $y_{n,k}$ is the actual value and ${\hat y}_{n,k}$ is the predicted value.

In this paper, data from the ISO\_NE Regional Transmission Organization public database has been used\footnote{Data can be downloaded from www.iso-ne.com}.  We run our experiments for the  region of North East Massachusetts (NEMASS) which corresponds to the Boston area, where only one weather station was needed to measure weather conditions, avoiding having to struggle with a large area where weather can vary.

Considering that several papers have found a strong relation between weather variables and load, hourly electricity demand data from 2011 until 2018 were used for this experiments as well as hourly temperature and dew point data from the Boston's weather base station. Years 2011 until 2017 were used for training purposes and 2018 as a testing set.

\subsection{Examples of data selection}

Here we present examples of data selection using the three methods. The data is selected based on test samples chosen from each one of the four seasons. For the CT method, The used data is from year 2016 and 2017, with a total of 714 samples. For the other methods, 714 samples of data have been selected from years 2011 to 2017. The four test samples used for the selection are chosen of months January, May, August and October. 

The input data to each of the selectors consists of the present day temperature, the next day forecast temperature and the present day hourly load. The details and dimensions of the data are in Table \ref{tab:pattern_dimensions1}.

\begin{table}[H]
\caption{Data format for training the data selection model. }
\centering
\begin{tabular}{|l|c|c|}
\hline
Name & Symbol & Dimension\\
\hline
Present day temperature & $\bT_i$ & $24$\\
 \hline
Next day (forecast) temperature & $\bT_{i+1}$ & $24$\\
 \hline
 Load & ${\bf l}_i$ & $24$ \\
\hline 
\end{tabular}
\label{tab:pattern_dimensions1}
\end{table}

\begin{figure}[H]
\centering
\includegraphics[width=6cm]{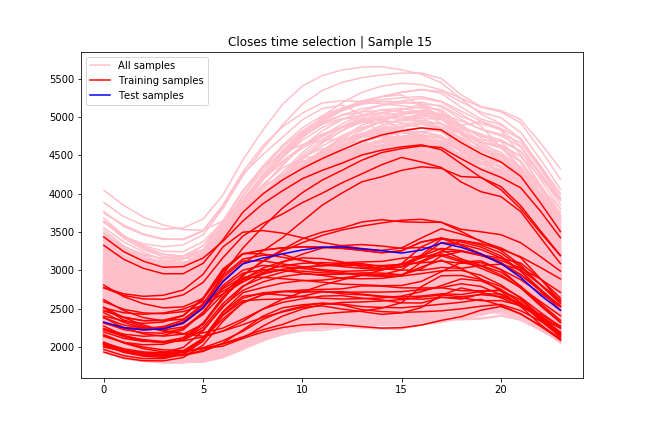}~\includegraphics[width=6.5cm]{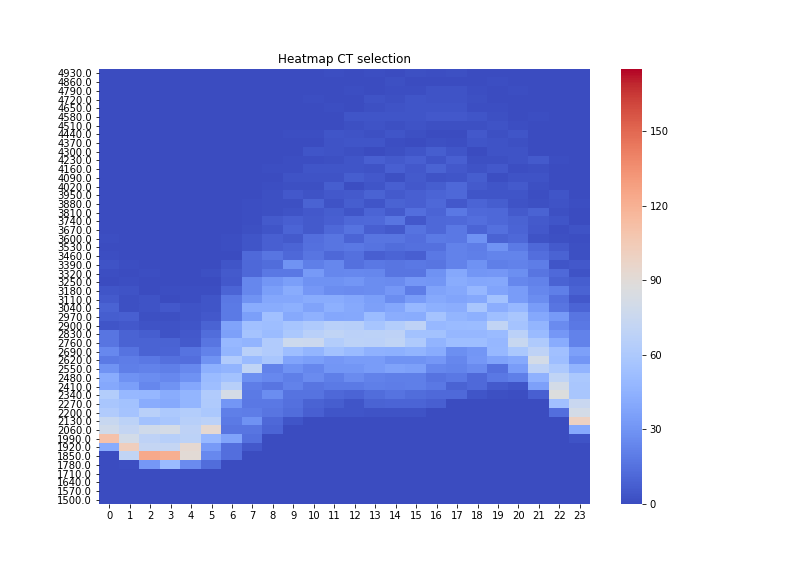}

\includegraphics[width=6cm]{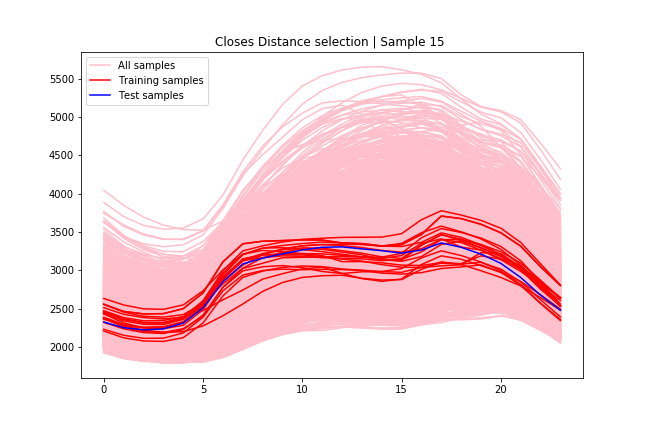}
\includegraphics[width=6.5cm]{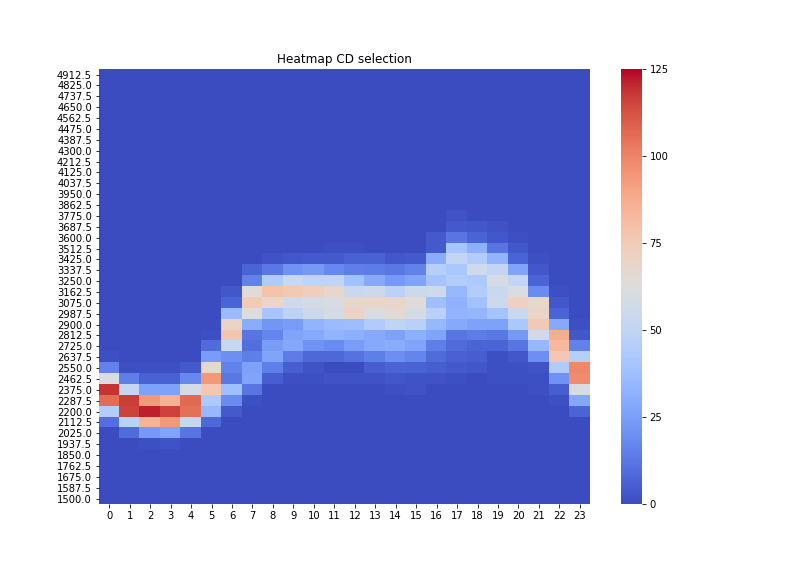}

\includegraphics[width=6cm]{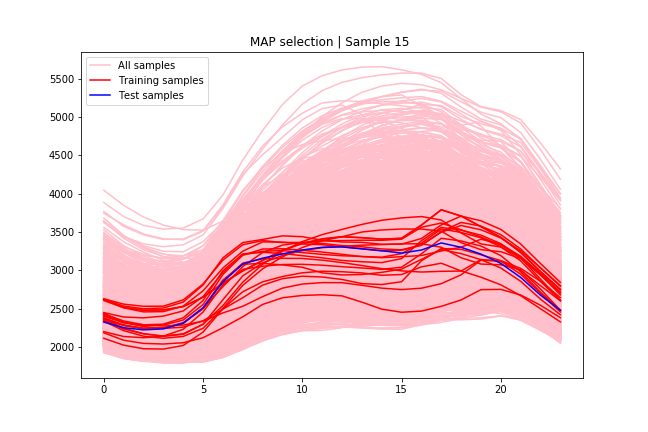}
\includegraphics[width=6.5cm]{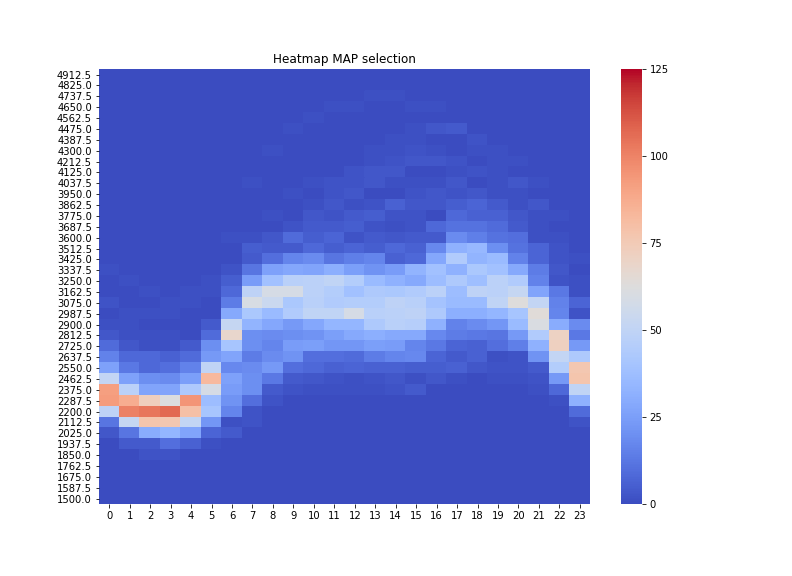}
\caption{Data selected using all three methods when the test sample is from January. The left column contains the data selected in red and the test samples shown in blue. The right column represents the density map of the selected time series as a function of time. }
\label{fig:SelectionWinter}
\end{figure}

Figures \ref{fig:SelectionWinter} and subsequent ones compare the data selected using CD and MAP method. Left panels show the similarity between samples selected and the test sample on each case. On the right panel, a density map shows how close samples are from the test one.

On the top panel, using CT data selection there is not an actual algorithm applied to it, but a raw data selection where all the samples from 2017 are selected forming a 357 samples training set. We can see that many of the samples do not have a  trend similar to the test sample, so they will not provide valid information during the training.On the middle panel, CD data selection selects samples that are very similar to out test sample, mainly from the same season as the test sample.  On the bottom panel, the MAP selection also chooses samples similar to the test samples that also provide a stronger importance for prediction.

\begin{figure}[H]
\centering

\includegraphics[width=6.5cm]{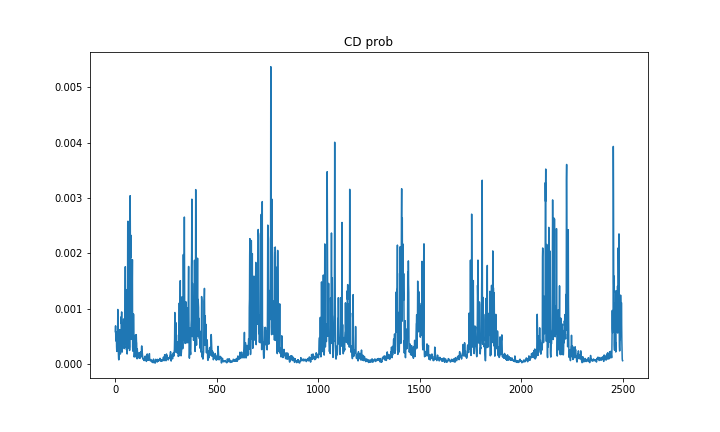}~
\includegraphics[width=6.5cm]{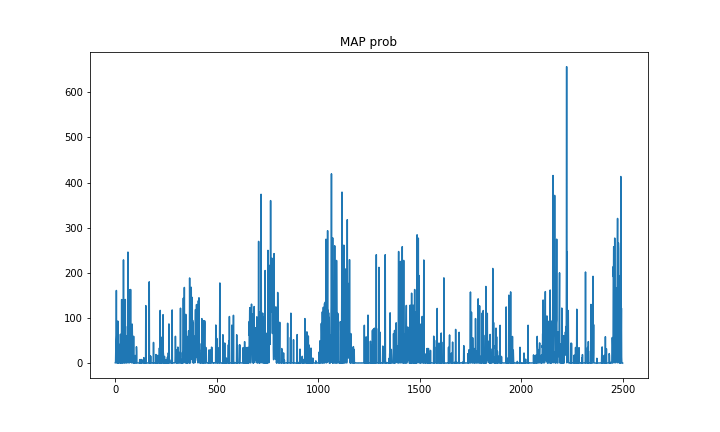}
\caption{Posterior probability density estimation computed using the CD (left) and MAP methods (right) when the test sample is from January. The CD method clearly captures the seasonal behaviour of the data, while the MAP method is less clear as it selects data from other seasons that have a similarity with the test data. }
\label{fig:posteriorWinter}
\end{figure}

Figures \ref{fig:posteriorWinter} and subsequent ones show the posterior probabilities of the time instants, that are here sorted from 2011 to 2017, for a winter 2018 test sample. The maxima correspond to  winter samples between 2011 and 2017. They have higher probabilities and therefore they are are selected to train the model.

\begin{figure}[H]
\centering
\includegraphics[width=6cm]{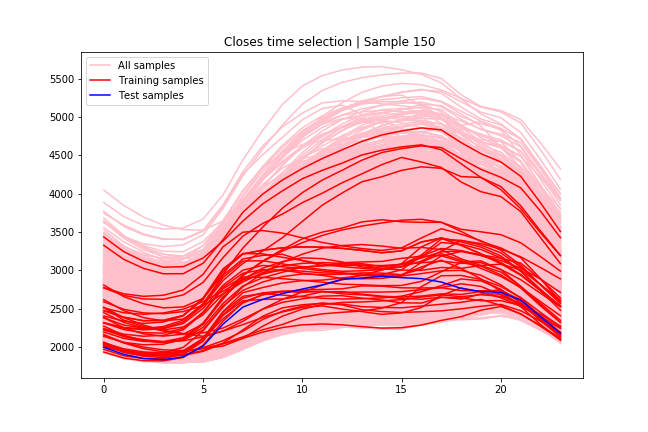}~\includegraphics[width=6.5cm]{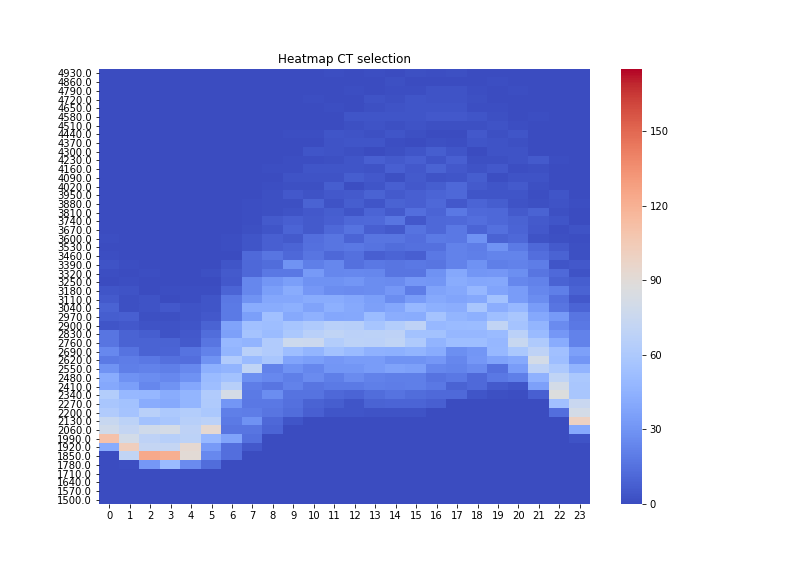}

\includegraphics[width=6cm]{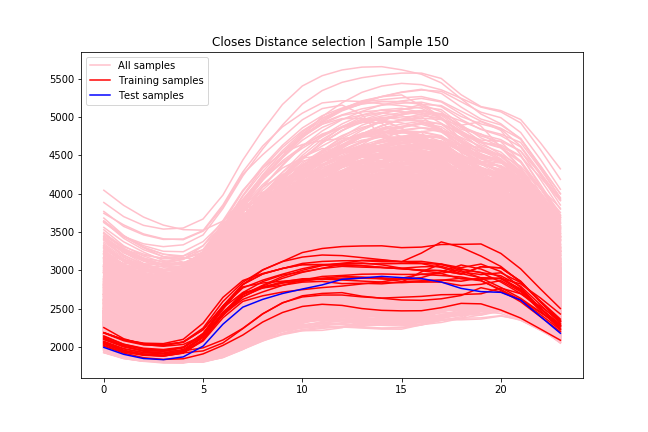}
\includegraphics[width=6.5cm]{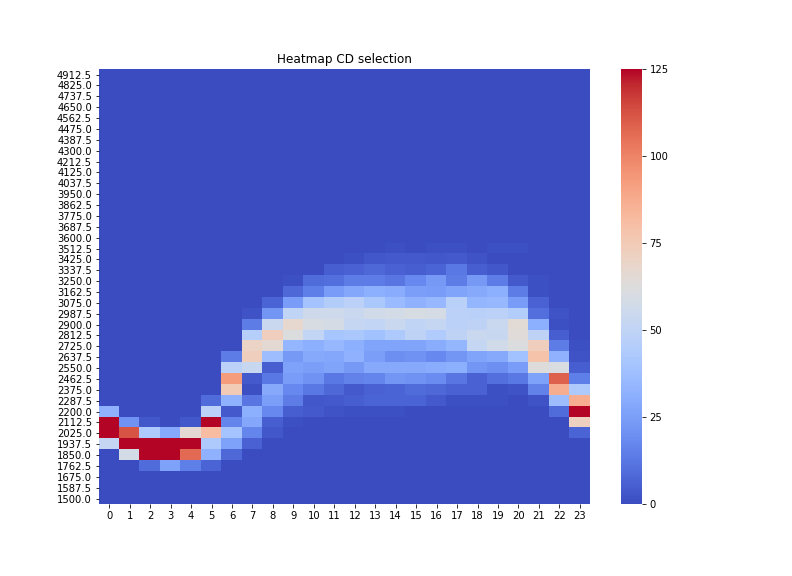}

\includegraphics[width=6cm]{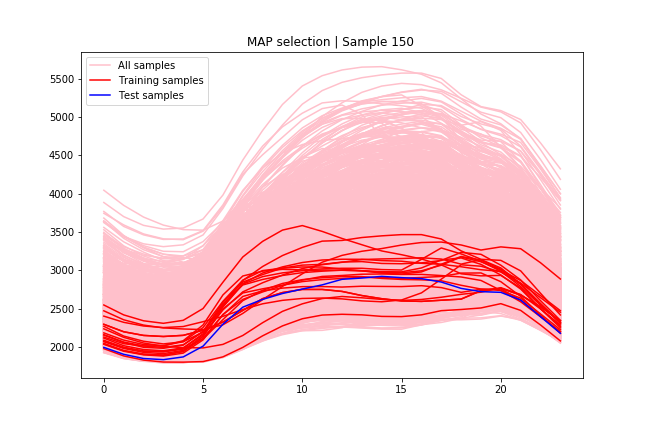}
\includegraphics[width=6.5cm]{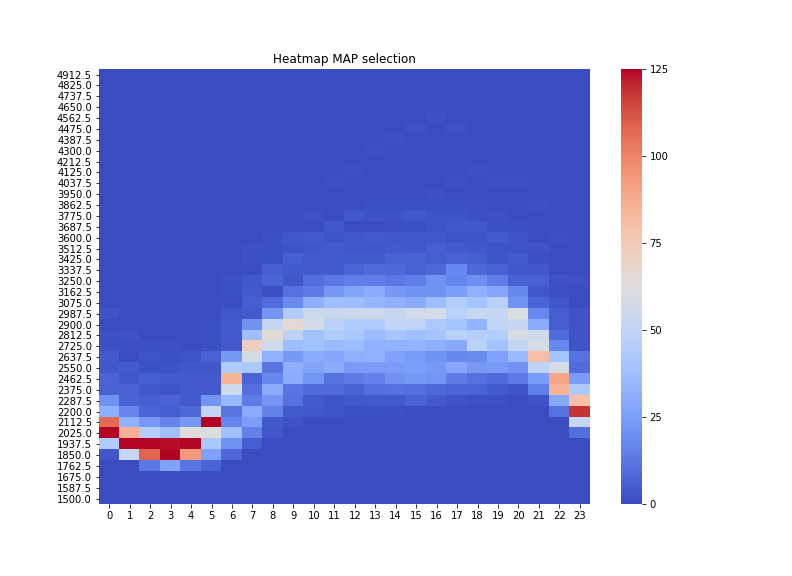}
\caption{Data selected using all three methods when the test sample is from May. The left column contains the data selected in red and the test samples shown in blue. The right column represents the density map of the selected time series as a function of time. }
\label{fig:SelectionSpring}
\end{figure}

\begin{figure}[H]
\centering

\includegraphics[width=6.5cm]{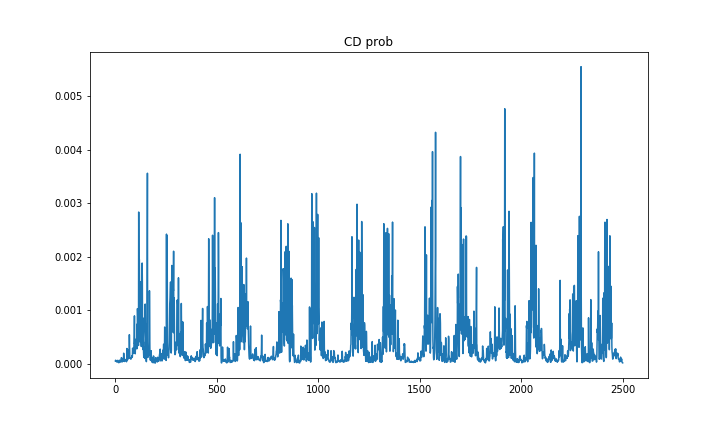}~
\includegraphics[width=6.5cm]{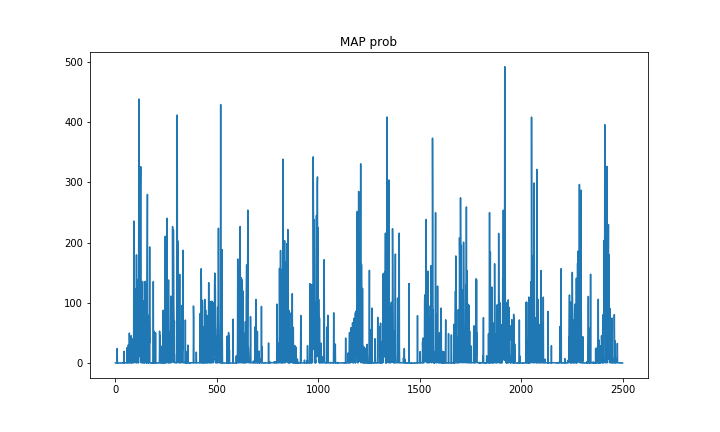}
\caption{Posterior probability density estimation computed using the CD (left) and MAP methods (right) when the test sample is from May. The CD method clearly captures the seasonal behaviour of the data, while the MAP method is less clear as it selects data from other seasons that have a similarity with the test data. }
\label{fig:posteriorSpring}
\end{figure}

An equivalent comparison is shown in Figure \ref{fig:SelectionSpring}, where a  spring day is selected as the test sample. We can see on figure \ref{fig:posteriorSpring} how higher probabilities are assigned to spring and fall days since there is a similarity between these two seasons. This is, then, a method that differs from a criteria that would strictly choose data of the same season in different years for training.

\begin{figure}[H]
\centering
\includegraphics[width=6cm]{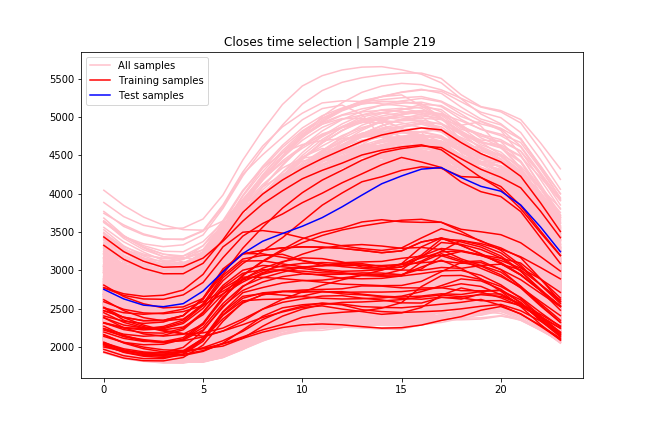}~\includegraphics[width=6.5cm]{pics/CT_heatmap_Winter.png}

\includegraphics[width=6cm]{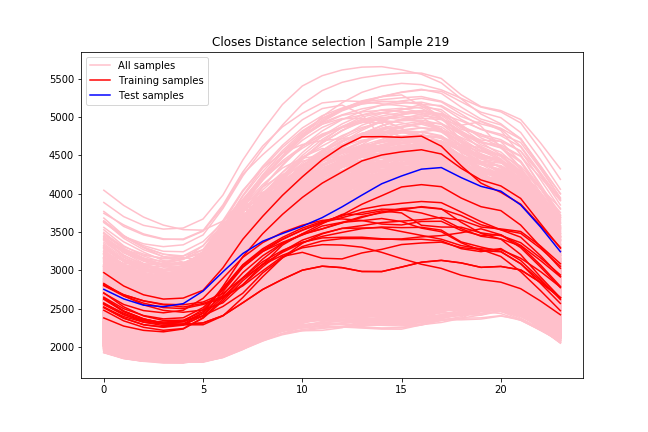}
\includegraphics[width=6.5cm]{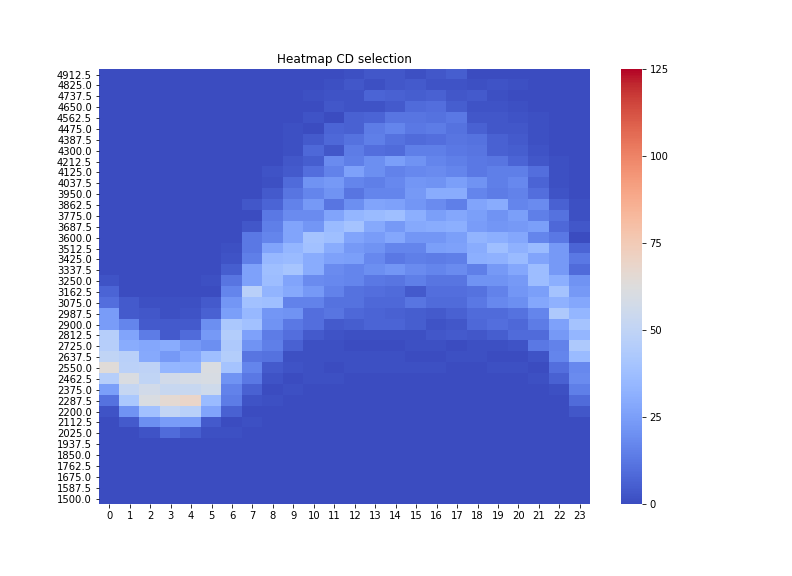}

\includegraphics[width=6cm]{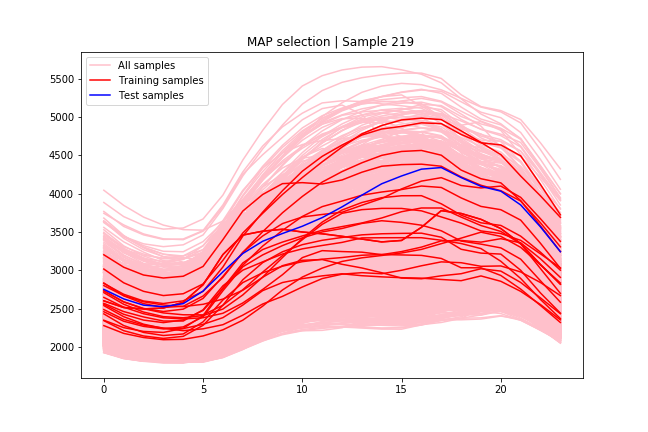}
\includegraphics[width=6.5cm]{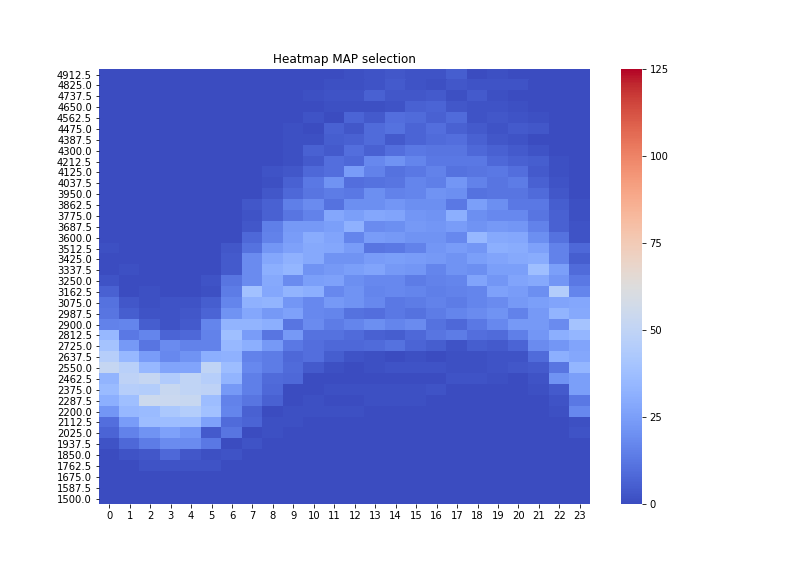}
\caption{Data selected using all three methods when the test sample is from July. The left column contains the data selected in red and the test samples shown in blue. The right column represents the density map of the selected time series as a function of time. }
\label{fig:SelectionSummer}
\end{figure}

\begin{figure}[H]
\centering

\includegraphics[width=6.5cm]{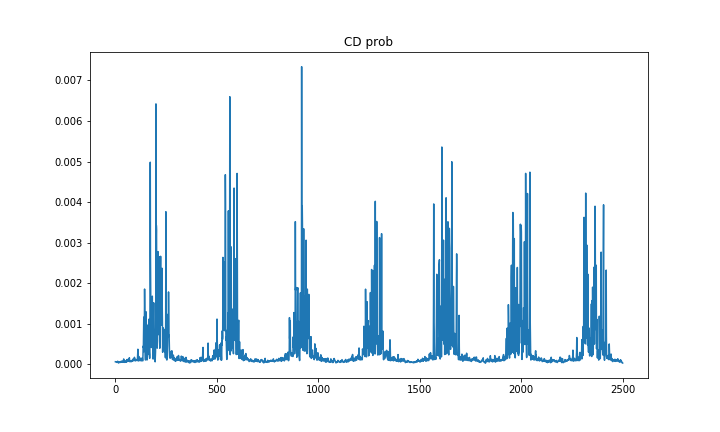}~
\includegraphics[width=6.5cm]{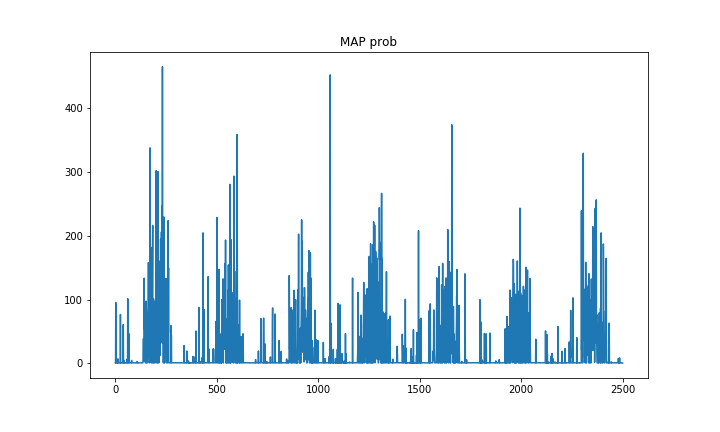}
\caption{Posterior probability density estimation computed using the CD (left) and MAP methods (right) when the test sample is from 
July. The CD method clearly captures the seasonal behaviour of the data, while the MAP method is less clear as it selects data from other seasons that have a similarity with the test data. }
\label{fig:posteriorSummer}
\end{figure}

In Figure \ref{fig:SelectionSummer}, a summer day is selected as the test sample. Just like in winter data selection, we can see on figure \ref{fig:posteriorSummer} how higher probabilities are assigned to mainly summer days. The MAP model is revealing here that a very few spring and fall days have similar behaviour and thus they can be used for the training of a summer forecast.

\begin{figure}[H]
\centering
\includegraphics[width=6cm]{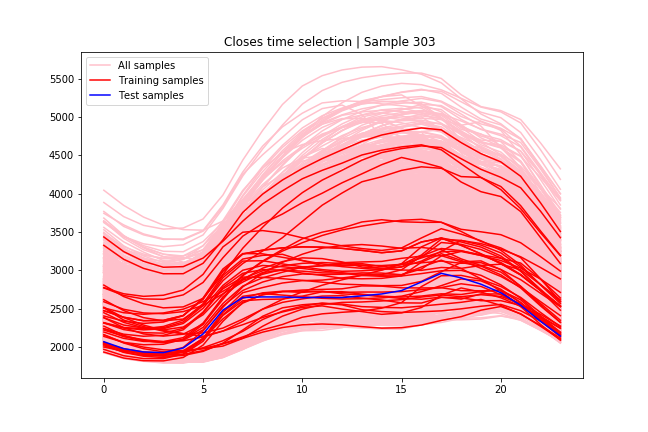}~\includegraphics[width=6.5cm]{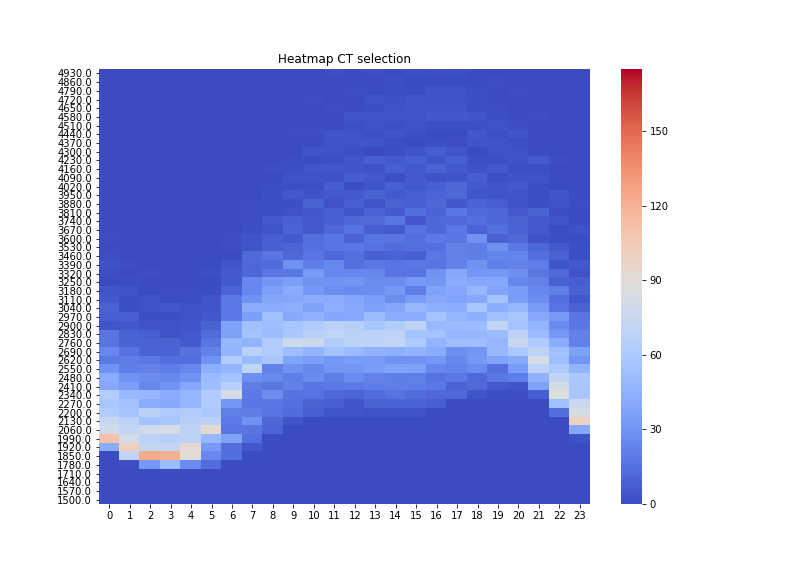}

\includegraphics[width=6cm]{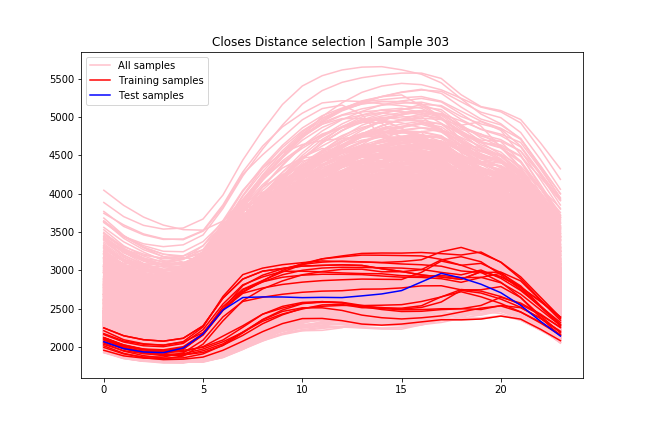}
\includegraphics[width=6.5cm]{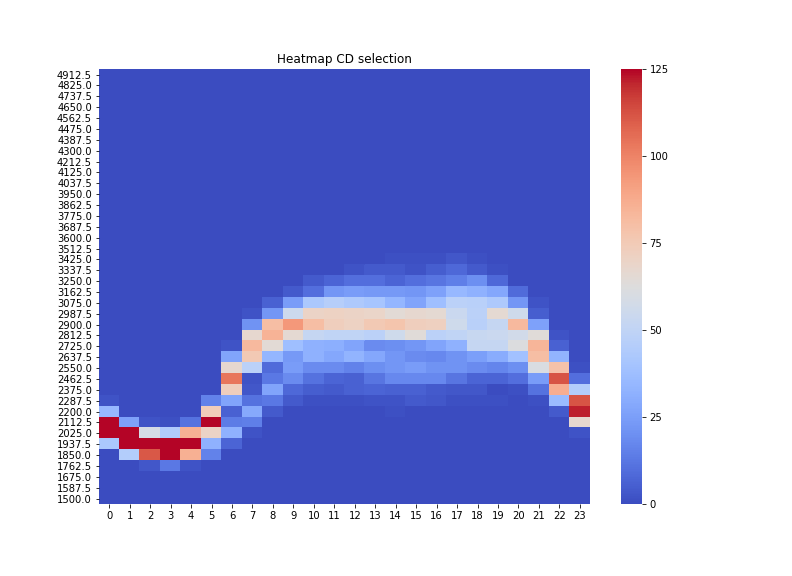}

\includegraphics[width=6cm]{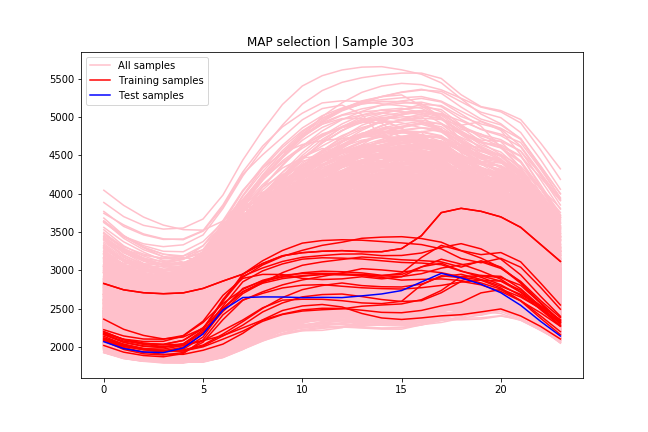}
\includegraphics[width=6.5cm]{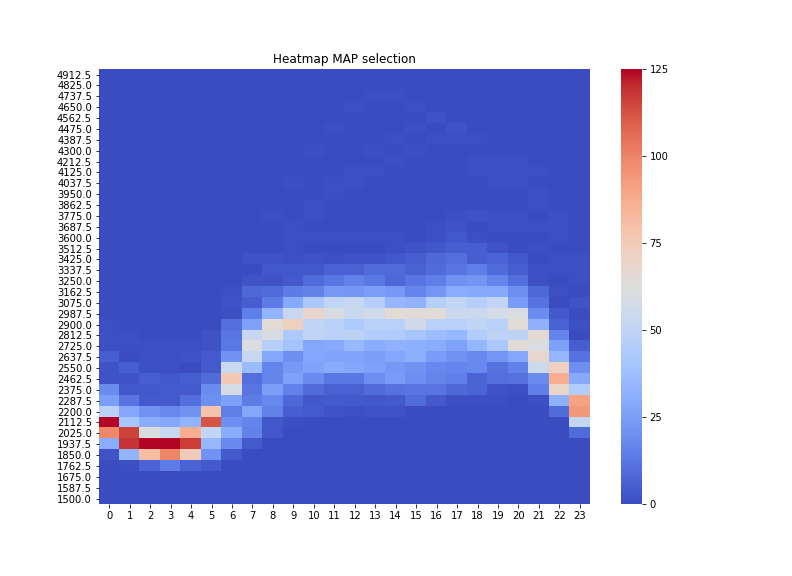}
\caption{Data selected using all three methods when the test sample is from November. The left column contains the data selected in red and the test samples shown in blue. The right column represents the density map of the selected time series as a function of time.}
\label{fig:SelectionFall}
\end{figure}

\begin{figure}[H]
\centering

\includegraphics[width=6.5cm]{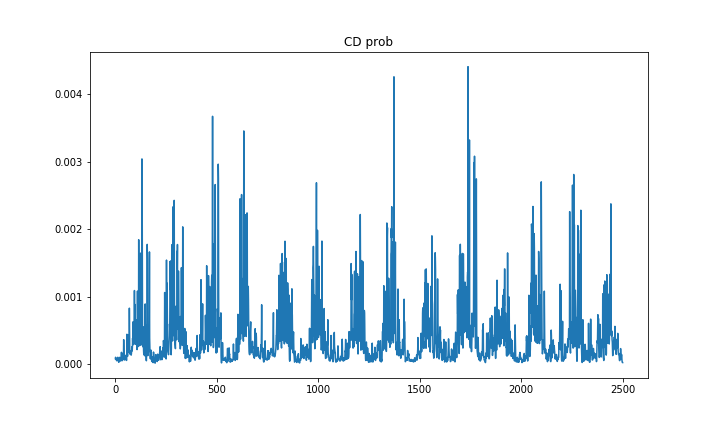}~
\includegraphics[width=6.5cm]{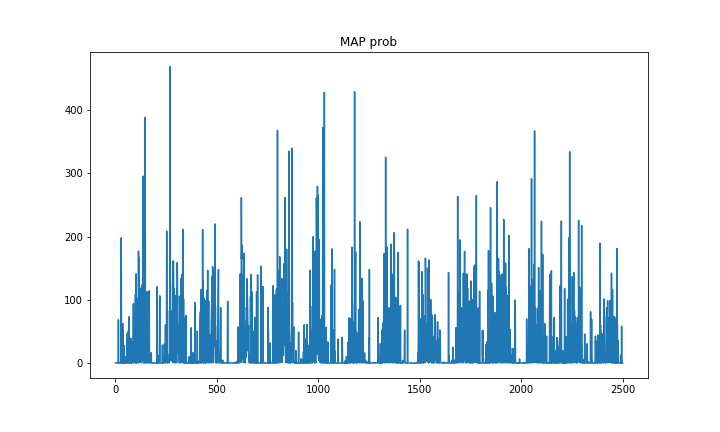}
\caption{Posterior probability density estimation computed using the CD (left) and MAP methods (right) when the test sample is from 
November. The CD method clearly captures the seasonal behaviour of the data, while the MAP method is less clear as it selects data from other seasons that have a similarity with the test data. }
\label{fig:posteriorFall}
\end{figure}

An equivalent comparison is shown in Figure \ref{fig:SelectionFall}, where a  fall day is selected as the test sample. We can see on figure \ref{fig:posteriorFall} how higher probabilities are assigned to spring and fall days. The model is revealing here that some spring and fall days have similar behaviour and thus they can be used for the training of a fall forecast. 

\subsection{One day ahead forecast with Gaussian Process regression}
 In this section we compare the introduced MAP selection method
in short term load forecast training and test against the CT and CD methods. The predictor forecasts one day ahead in 24 points spaced at 1 hour intervals. The structure is simply a set of 24 independent linear Gaussian Process regressors \cite{Rasmussen:2006}. The test data are all days of 2018 and the training data are days selected between 2011 to 2017. The result of each test is a set of 24 Gaussian predictive posterior distributions, characterized by the predicted mean and variance. With this, a confidence interval is constructed to estimate the quality of the prediction. 

The input data to each of the predictors consists of the time stamp, the past week's daily temperature, past week's daily load, the present day dew point and temperature, the temperature and dew point forecasted for the next day and the present day hourly load. The details and dimensions of the data are in Table \ref{tab:pattern_dimensions}.

\begin{table}[h!]
\caption{Data used as predictor for the 24 hour ahead prediction. }
    \centering
    \begin{tabular}{|l|c|c|}
        \hline
        Name & Symbol & Dimension\\
        \hline
         Time stamp & $t_{i}$ & $1$\\
         \hline
         Past week's temperatures &  $\bT_{i-6:i-1}$ & $144$\\
         \hline
         Past week's loads &  $\bL_{i-6:i-1}$ & $144$\\
         \hline
         Present day dew point  & $\bdp_i$ & $24$\\
         \hline
         Next day's (forecast) dew point & $\bdp_{i+1}$ & $24$\\
         \hline
         Present day temperature & $\btemp_i$ & $24$\\
         \hline
         Next day (forecast) temperature & $\btemp_{i+1}$ & $24$\\
         \hline
         Present day load & ${\bf l}_i$ & $24$ \\
        \hline 
    \end{tabular}
    \label{tab:pattern_dimensions}
\end{table}


Table \ref{tab:forecasting_baseline}  shows some forecasting results using several regression methods for day ahead hourly load forecasting are shown where ISO-NE dataset or similar american utility and transmission companies were used.

\begin{table}[H]
\caption{Other methods using American utility companies datasets}
\centering
\begin{tabular}{|c|c|}
\hline
Method & MAPE \\
\hline
LW-SVR\footnotemark[1] \cite{Ehab:2010} & 3.62 \\
\hline
local SVR\footnotemark[2]  \cite{Ehab:2010} & 4.08\\
\hline
LWR\footnotemark[3]  \cite{Ehab:2010} & 4.71\\
\hline
MLR\footnotemark[4]  \cite{Xie:2018} & 3.63\\
\hline
Hong\footnotemark[5] \cite{Feng:2016} & 3.75\\
\hline
TWE\footnotemark[6] \cite{Feng:2016} & 3.25\\
\hline
CS\footnotemark[7] \cite{Charlton:2014} & 3.5\\
\hline
\end{tabular}
\label{tab:forecasting_baseline}
\end{table}

\footnotetext[1]{Locally Weighted Support Vector Regression}
\footnotetext[2]{Local Support Vector Regression}
\footnotetext[3]{Locally Weighted Regression}
\footnotetext[4]{Multiple Linear Regression}
\footnotetext[5]{T.Hong Ph.D. thesis}
\footnotetext[6]{Temporal and Weather conditional epi-splines load model}
\footnotetext[7]{refined parametric model for STLF}

Table \ref{tab:forecasting_GP} shows the results of a Gaussian Process regressor using all the different data selection methods. The MAPE of the MAP selection method is the best one and it outperforms in more than 
$0.5$ the results with the other two methods. Note that the MAPE is competitive with the state of the art methods shown in Table \ref{tab:forecasting_baseline}. 
Also, the MAP predicted variance is pretty similar to the CD variance. In order to test the accuracy of the variance, the number of predictions inside the confidence interval of $2\sigma$ was counted. The correct number of samples should be $95\%$, and the real one is $94.73\%$ as it is shown in Table \ref{tab:forecasting_GP} .

\begin{table}[H]
\caption{Performance comparisons across data selection methods for Gaussian Processes}
\centering
\begin{tabular}{|c|c|c|c|c|c|}
\hline
Method & MAPE & R2 score &  $\sigma^2_{pred}$ & 95{\%} & Data selection plus training time (s) \\
\hline
Closest Time Selection  & 2.67

 & 0.91
 & 6626
 & 93.62
 & 0.13 \\
 \hline
Closest Distance Selection & 2.1
 & 0.94
 & 3507
 & 90.1
 & 62 \\
 \hline
 {MAP Selection} & 2.02
 & 0.94
 & 3992
 & 94.73
 & 165 \\
 \hline
\end{tabular}
\label{tab:forecasting_GP}
\end{table}

On figure 9, a load prediction trained with the different sample selection is shown. Real value and predicted value are shown and compared and a 95{\%} confidence interval is added to show how confident and accurate are our prediction. 
The right panel shows a slightly wider estimated confidence interval, while the left panel has a confidence interval which is slightly smaller than the right one, but as shown in  Table \ref{tab:forecasting_GP}, a higher percentage of samples are in the MAP selection slightly bigger confidence interval. 

\begin{figure}[H]
\centering
\includegraphics[width=9cm]{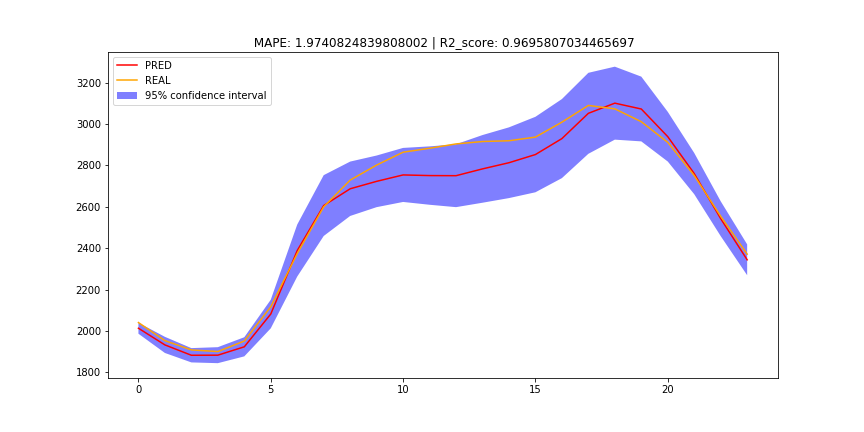}~\includegraphics[width=9cm]{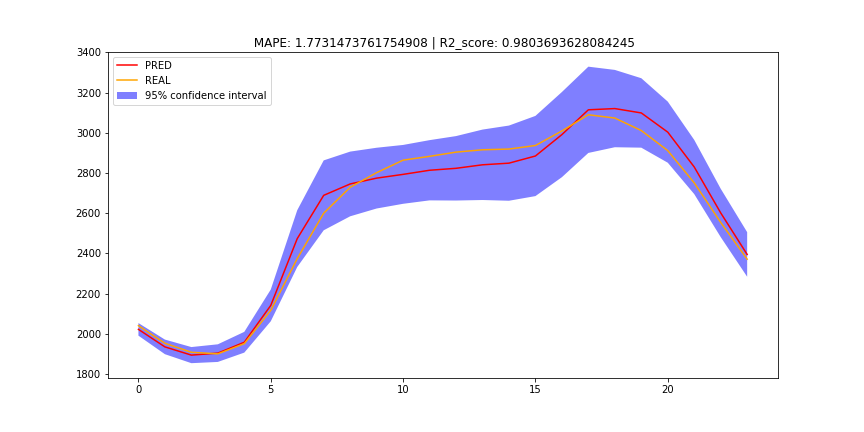}

\caption{Example of mean and confidence interval of a forecast. In the left panel, the data has been selected using the distance method, and in the right panel, the MAP method has been used. MAPE and R2 score are better on the right panel where MAP selection is applied. }
\label{simulationfigure17}
\end{figure}
\section{Conclusion}
This paper introduces a new method for data selection applied to power load forecast. The selection method is particularly tailored to process multidimensional time series, so it takes advantage of the fact that every sample of the data used for training and test has a timestamp that determines what day of the year the data was produced. The presented methodology is intended to compute a posterior probability of every time instant given the test data used to predict the load for the next day. This is, if the prior probability mass function to pick a day to include it in a training set is uniform, the posterior is modified by the similarity of the present day with the resat of the available data. 

In order to construct this posterior, a multivariate Gaussian likelihood function is constructed and maximized in a recursive way with the available data, and then, using the Bayes' theorem, a posterior of the timestamp is constructed. 

The algorithm has been tested with load databases between 2011 and 2017. If the test data belongs to Summer, for example, the algorithm will choose data from other summers, but it will exclude the anomalous data of these summers. If the test data is from Spring, then the algorithm naturally chooses data from Spring and also Fall, which is intuitively reasonable. 

The training data has been tested with Gaussian Process based predictors to show that the data selection significanlty improves its performance. 

The presented algorithm is restricted to linear models of the data, but the formulation allows the use of Mercer's kernels to capture the nonlinear behaviour of the data. Future work includes the use of nonlinear versions of the present algorithm.

\begin{appendices}
\section{Mean and covariance matrix of the posterior of $\bw$ }\label{app:matrix_inversion}

In this appendix we provide the proofs of equations  \eqref{eq:mean_weights} and \eqref{eq:covar_weights}. Assume a Gaussian prior on a set of weights $\bw$ and a Gaussian likelihood on a set of observations with the forms
\begin{equation}
    p(\bw)=\frac{1}{\sqrt{(2\pi)^D |\bSigma_p|}} 
    \exp 
    \left(
        \frac{1}{2}\bw^{\top}\bSigma_p^{-1}\bw
    \right)
\end{equation}
and 
\begin{equation}
            p(\by|\bX,\bw)=\frac{1}{\sqrt{2\pi} \sigma^D} \exp
            \left(
                \frac{1}{2\sigma^2}
                \left(
                    \by -\bw^{\top}\bX
                \right)^{\top}
                \left(
                    \by -\bw^{\top}\bX
                \right)
            \right)
\end{equation}
The posterior of $\bw$ given $\bX,\by$ can be easily computed using the Bayes' rule
\begin{equation}\label{eq:Bayes}
    p(\bw|\bX,\by)=\frac{p(\bw)p(\by|\bX,\bw)}{p(\by|\bX)}\propto p(\bw)p(\by|\bX,\bw)
\end{equation}
Since the posterior is proportional to the product of two Gaussians, it must be a Gaussian where its exponent has the following argument   
\begin{equation}\label{eq:exp_argument}
\begin{split}
&-\frac{1}{2\sigma^2}\left(
    \by -\bw^{\top}\bX
\right)^{\top}
\left(
    \by -\bw^{\top}\bX
    \right)-
    \frac{1}{2}\bw^{\top}\bSigma_p^{-1}\bw=\\
&-\frac{1}{2\sigma^2}\left(\by^{\top}\by +\bw^{\top}\bX\bX^{\top}\bw-2\by^{\top}\bX^{\top}\bw\right)-\frac{1}{2}\bw^{\top}\bSigma_p^{-1}\bw=\\
&-\frac{1}{2\sigma^2}\by^{\top}\by -\frac{1}{2\sigma^2}\bw^{\top}\left(\bX\bX^{\top}+\bsigma^2\bSigma_p^{-1}\right)\bw+\frac{1}{\sigma^2}\by^{\top}\bX^{\top}\bw\\
\end{split}
\end{equation}

If the posterior \eqref{eq:Bayes} is a Gaussian, it is proportional to an exponential whose argument is 
\begin{equation}
    -\frac{1}{2}\left(\bw -\bar{\bw}\right)^{\top}\bA\left(\bw -\bar{\bw}\right)
\end{equation}

By developing the above expression and comparing it to last line of equations \eqref{eq:exp_argument} we see that \begin{equation}\label{eq:pos_covar}
    \bA=\sigma^{-2}\bX\bX^{\top}+\bSigma^{-1}_p
    \end{equation}
and
that $\bar{\bw}^{\top}\bA=\sigma^{-2}\by^{\top}\bX^{\top}$,  hence 
\begin{equation} \label{eq:pos_mean}
\bar{\bw}=\sigma^{-2}\bA^{-1}\bX\by
\end{equation}

Now we identify 
\begin{equation}
\begin{split}
&\bx = \bx_d\\
&\bX =
    \left(
        \begin{array}{c}
            \bX_{1:d-1} \\
            \bt^{\top}
    \end{array}
    \right)\\
&{\bw} =
    \left(
        \begin{array}{c}
            \bw_d\\
            \bw_{t,d} 
        \end{array}
    \right)
\end{split}
\end{equation}
with $\sigma^2 = \sigma^2_d$, $\bSigma_p=\bSigma_{p,d}$, and then $\bar{\bw}$ and $\bA$ become

\begin{equation}
\bar{\bw}=\left(
\begin{array}{c}
\bar{\bw}_d\\
\bar{\bw}_{t,d}
\end{array}
\right)=\sigma_d^{-2}\bA_d^{-1}\left(
\begin{array}{c}
\bX_{1:d-1}\\
\bt^{\top}
\end{array}\right)\bx_d
\end{equation}
and 
\begin{equation}
\bA=\bA_d=\sigma^{-2}_d\left(
\begin{array}{cc}
\bX_{1:d-1}\bX_{1:d-1}^{\top} &\bX_{1:d-1}\bt^{\top}\\
\bt\bX_{1:d-1}^{\top}&\sum_n t_{n}^2
\end{array}
\right)+\bSigma_{p,d}^{-1}
\end{equation}

\section{Predictive posterior derivation}\label{app:posterior}

In this appendix we provide the proof of equation \eqref{eq:posterior_variance}. This result, as the one presented in Appendix A is very well, known, but seldom proved in the literature, and this is why we present it here for the interested reader. 

Assume, with the same notation and reasoning as in  \cite{Rasmussen:2006} the output $f(\bx_*)=f_*$ of a regression model for sample $\bx_o$

The predictive posterior is found by computing the expectation of the distribution of the output of all possible models across the parameters $\bw$, this is

\begin{equation}
    p(f_*|\bX,\by,\bx^*)=\int_{\bw} p(f_*|\bw,\bx^*)p(\bw|\bX,\by)d\bw = \int_{\bw} p(f_*,\bw,|\bx^*,\bX,\by)d\bw
\end{equation}
where $p(f_*|\bw,\bx^*)=\mathcal{N}(f_*|\bw^{\top}\bx,\sigma^2)$ and $p(\bw|\bX,\by)=\mathcal{N}(\bw|\bar{\bw},\bA)$, with mean and covariance matrices given by equations  \eqref{eq:pos_mean} and \eqref{eq:pos_covar} and where the integral is simply the marginalization over $\bf w$ of the joint probability distribution of $\bf w$ and $f_*$. Then, the mean of $f_*$ with respect to $p(f_*|\bX,\by,\bx^*)$ is 
\begin{equation}
    \mathbb{E}(f_*)=\mathbb{E}_{\bw}(\bw^{\top}\bx^*)={\bar\bw}^{\top}\bx^*=\sigma^{-2}\by^{\top}{\bA}^{-1}{\bx}^*
\end{equation}
and the variance is 
\begin{equation}
    \mathbb{V}ar(f_*)=\mathbb{V}ar_{\bw}(\bw^{\top}\bx^*)=\mathbb{E}_{\bw}\left((\bw^{\top}\bx^*)^2\right)-\mathbb{E}_{\bw}^2(\bw^{\top}\bx^*)=\bx^{\top}\mathbb{E}_{\bw}(\bw\bw^{\top})\bx=\bx^{\top}\bA^{-1}\bx
\end{equation}

\section{Recursive computation of inverse matrices}\label{app:recursive}

Assuming that $\bSigma_{p}=\bI$  we can rewrite matrix $\bA_{d}$  as

\begin{equation}
\bA_{d}=\left(
\begin{array}{ccc}
|\bx_{d-1}|^2&
\bx_{d-1}\bX_{d-2}^{\top}&\bx_{d-1}\bt^{\top}\\
\bX_{d-2}\bx_{d-1}^{\top}&\bX_{d-2}\bX_{d-2}^{\top}&\bX_{d-2}\bt^{\top}\\
\bt\bx_{d-1}^{\top}& \bt\bX^{\top}_{t-2}&\bt\bt^{\top} 
\end{array}
\right)+\sigma^{2}_t\bI
\end{equation}
where vector $\bx_{d-1}=[x_{1,d-1}\cdots x_{N,d-1}]$ is just the first row of matrix $\bX_{d-1}$
This is, matrix $\bA_d$ is constructed by adding a row and a column to the previous matrix. Then, one can rewrite this matrix in a recursive way as 
\begin{equation}
\bA_{d}=\left(
\begin{array}{cc}
	|\bx_{d-1}|^2+\sigma^{2}_t&
	\begin{array}{cc}
			\bx_{d-1}\bX_{d-2}^{\top} & \bx_{d-1}\bt^{\top}
	\end{array}\\
	\begin{array}{c}
		\bX_{d-2}\bx_{d-1}^{\top}\\
		\bt\bx_{d-1}^{\top}
	\end{array}
& \bA_{d-1} 
\end{array}
\right)
\end{equation}
where the first element of the recursion is $\bA_1=\bX\bX^{\top}+\sigma^2_1\bI$. We now assume that matrix $\bA^{-1}_{d-1}$ is known. Then, $\bA_{d}^{-1}$ can be computed by recursion. We define it in four blocks as
\begin{equation}
\bA^{-1}_{d}=\left(
\begin{array}{cc}
a&\bb\\
\bb^{\top}&\bC
\end{array}
\right)
\end{equation}
Using the matrix inversion lemmas
\begin{eqnarray}
&a=&\left(|\bx_{d-1}|^2+\sigma^2_d-\bu^{\top}\bA^{-1}_{d-1}\bu\right)^{-1}\\
&\bb=&-(|\bx_{d-1}|^2+\sigma^2_d)\bu\bC\\
&\bC=&\left(\bA^{-1}_{d-1}+(|\bx_{d-1}|^2+\sigma^2_d)\frac{\bA^{-1}_{d-1}\bu\bu^{\top}\bA^{-1}_{d-1}}{\sigma^2_d-\bu\bA^{-1}_{d-1}\bu}\right)\\
&\bu=&\left[\bX_{d-2}^{\top}~~~\bt^{\top}\right]\bx_{d-1}^{\top}
\end{eqnarray}

\end{appendices}
\section*{Acknowledgements}
This work is based upon work supported in part by the National Science Foundation EPSCoR Cooperative Agreement OIA-1757207 and TEC 2014-52289-R. 

Authors would like to thank UNM Center for Advanced Research Computing, supported in part by the NSF, for providing high performance computing, large-scale storage and visualization resources.

\bibliographystyle{IEEEtran}
\bibliography{references}

\end{document}